\documentclass[letterpaper]{article} % DO NOT CHANGE THIS
\usepackage{aaai25}  % DO NOT CHANGE THIS
\usepackage{times}  % DO NOT CHANGE THIS
\usepackage{helvet}  % DO NOT CHANGE THIS
\usepackage{courier}  % DO NOT CHANGE THIS
\usepackage[hyphens]{url}  % DO NOT CHANGE THIS
\usepackage{graphicx} % DO NOT CHANGE THIS
\usepackage{natbib}  % DO NOT CHANGE THIS AND DO NOT ADD ANY OPTIONS TO IT
\usepackage{enumitem}
\usepackage{amssymb}
\usepackage{caption} % DO NOT CHANGE THIS AND DO NOT ADD ANY OPTIONS TO IT
\usepackage{algorithm}
\usepackage{algorithmic}
\usepackage{amsmath}
\usepackage[most]{tcolorbox}
\newtcolorbox{promptbox}{
    enhanced,
    colback=gray!5,
    colframe=gray!50,
    boxrule=0.5pt,
    fontupper=\small,
    arc=2pt,
    outer arc=2pt,
    boxsep=5pt,
    left=8pt,
    right=8pt,
    top=6pt,
    bottom=6pt
}
\usepackage{newfloat}
\usepackage{listings}
\DeclareCaptionStyle{ruled}{labelfont=normalfont,labelsep=colon,strut=off} % DO NOT CHANGE THIS
\floatstyle{ruled}
\newfloat{listing}{tb}{lst}{}
\floatname{listing}{Listing}
\title{Domain-specific Question Answering with Hybrid Search}
\author{
    Dewang Sultania \equalcontrib, 
    Zhaoyu Lu \equalcontrib,
    Twisha Naik \equalcontrib,
    Franck Dernoncourt, \equalcontrib \\
    Seunghyun Yoon \equalcontrib,
    Sanat Sharma,
    Trung Bui,
    Ashok Gupta, \\
    Tushar Vatsa,
    Suhas Suresha,
    Ishita Verma, 
    Vibha Belavadi,
    Cheng Chen,
    Michael Friedrich
}
\affiliations{
    Adobe Inc.\\
    345 Park Avenue\\
    San Jose, CA, 95110\\
}

\usepackage{bibentry}
\begin{document}

\maketitle

\begin{abstract}

Domain-specific question answering is an evolving field that requires specialized solutions to address unique challenges. In this paper, we show that a hybrid approach—combining a fine-tuned dense retriever with keyword-based sparse search methods—significantly enhances performance. Our system leverages a linear combination of relevance signals, including cosine similarity from dense retrieval, BM25 scores, and URL host matching, each with tunable boost parameters. Experimental results indicate that this hybrid method outperforms our single-retriever system, achieving improved accuracy while maintaining robust contextual grounding. These findings suggest that integrating multiple retrieval methodologies with weighted scoring effectively addresses the complexities of domain-specific question answering in enterprise settings.

\end{abstract}


\section{Introduction}
With the increasing adoption of Large Language Models (LLMs) in enterprise settings, ensuring accurate and reliable question-answering systems remains a critical challenge. Building upon our previous work on domain-specific question answering about Adobe products  \cite{sharma2024retrieval}, which established a retrieval-aware framework with self-supervised training, we now present a production-ready, generalizable architecture alongside a comprehensive evaluation methodology.
Our core contribution is a flexible, scalable framework built on Elasticsearch that can be adapted for any LLM-based question-answering system. This framework seamlessly integrates hybrid retrieval mechanisms, combining dense and sparse search with boost matching, while maintaining production-grade performance requirements. While we demonstrate its effectiveness using our organization's domain-specific data, the architecture is designed to be domain-agnostic and can be readily deployed for other enterprise applications.
We evaluate our system through a rigorous methodology that assesses performance across multiple dimensions: context relevance through normalized Discounted Cumulative Gain (nDCG), response groundedness, answer accuracy, and answer null rate. To ensure robust testing, we compile a diverse evaluation dataset including:
\begin{enumerate}
\item Human-annotated responses for common feature queries
\item A carefully curated ``negative'' dataset containing jailbreak attempts, NSFW content, and irrelevant queries to test system boundaries
\item LLM-based comparative analysis between system outputs and human-annotated ground truth
\end{enumerate}
This comprehensive evaluation approach allows us to assess not only the system's ability to provide accurate information but also its robustness against inappropriate queries and its capability to acknowledge when information is not available. Our hybrid search mechanism demonstrates significant improvements across these metrics compared to single-method approaches.
The key contributions of this work include:
\begin{itemize}
\item A production-ready, generalizable framework for LLM-based QA systems built on Elasticsearch
\item A flexible hybrid retrieval mechanism combining dense and sparse search methods
\item A comprehensive evaluation framework for assessing QA system performance
\item Empirical analysis demonstrating the effectiveness of our approach across various metrics
\end{itemize}
Through this work, we provide not only theoretical insights but also a practical, deployable solution for building reliable domain-specific question-answering systems that can be adapted to various enterprise needs.

\section{Related Work}

Our research builds upon recent advancements in retrieval-augmented language models, domain-specific question answering (QA), and hybrid retrieval methods. We review relevant literature across these key areas.

\subsection{Retrieval-Augmented Language Models}

Retrieval-Augmented Generation (RAG) models integrate information retrieval with language models to enhance performance on knowledge-intensive tasks such as question answering. These models employ a retriever to identify pertinent documents, conditioning the language model on the retrieved context during response generation. \citet{lewis2020retrieval} introduced the RAG framework, demonstrating its effectiveness in open-domain QA by combining dense passage retrieval with sequence-to-sequence generation. Further, \citet{lazaridou2022internet} explored the application of RAG in an enterprise customer support setting, highlighting its potential for domain-specific QA.

\subsection{Domain-Specific Question Answering}

Developing QA systems tailored to specialized domains necessitates techniques that effectively capture domain-specific knowledge and terminology. \citet{sharma2024retrieval} proposed a framework for compiling domain-specific QA datasets and introduced retrieval-aware fine-tuning of language models, which reduces hallucinations and enhances contextual grounding. \citet{eppalapally-etal-2024-kapqa} introduced a model for query rewriting to take the domain of interest into account. Additionally, hybrid retrieval methods that combine dense and sparse techniques have shown promise in domain-specific QA. For instance, \citet{zhu2023hybrid} presented a hybrid text generation-based query expansion method, improving retrieval accuracy and QA system performance.

\subsection{Hybrid Retrieval Methods}

Integrating multiple retrieval signals can improve ranking performance compared to single-method approaches.  \citet{arivazhagan2023hybrid} explored hybrid hierarchical retrieval, combining sparse and dense methods in a two-stage document and passage retrieval setup, demonstrating improved in-domain and zero-shot generalization. Similarly, \citet{li2021dual} proposed a dual reader-parser architecture that leverages both textual and tabular evidence, enhancing open-domain QA performance.

While prior work has made significant strides, developing reliable and robust domain-specific QA systems for enterprise settings remains an open challenge. Our work addresses this by proposing a flexible, generalizable framework that integrates hybrid retrieval with weighted relevance scoring. We extend prior hybrid approaches with a multi-phase scoring algorithm and introduce a comprehensive evaluation methodology to assess system performance across multiple dimensions.

\section{Methodology}

To optimize document retrieval accuracy, we employed a multi-phase scoring algorithm that evolved through iterations, starting from a baseline BM25 model, progressing to a chunking-based approach, and finally integrating host-based score adjustments.

\begin{figure*}[!t]
    \centering
    \includegraphics[width=\textwidth]{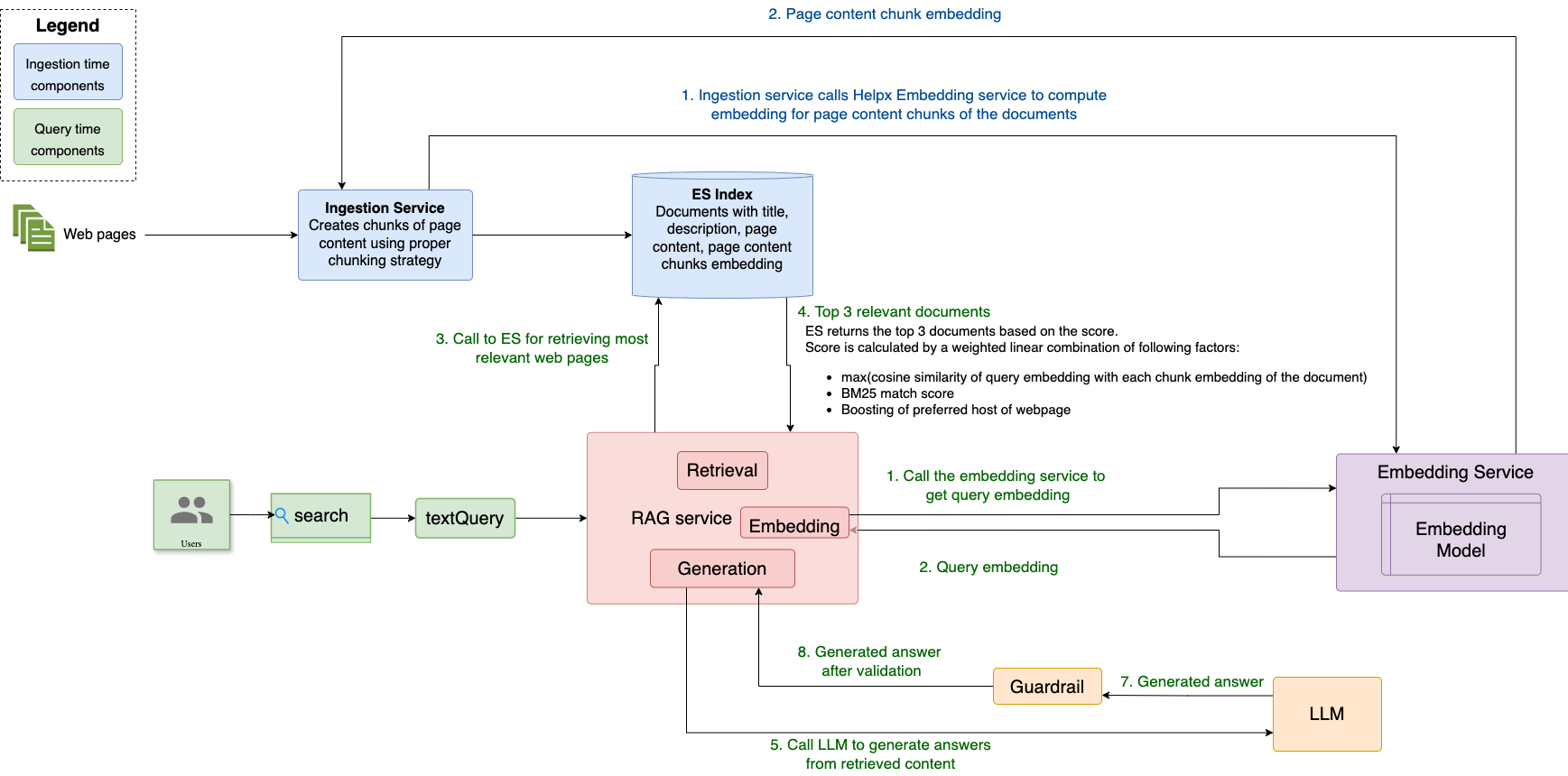}
    \caption{Production ready RAG deployment}
    \label{fig:architecture}
\end{figure*}

\subsection{Scoring Algorithm}

Given a user query, we calculate relevance scores for each document as follows:

\begin{align}
\text{score} = &\ \max(\text{matched chunks cosine score}) \notag \\
               & + \text{bm25\_boost} \times \text{BM25 score} \notag \\
               & + \text{host\_boost} \times \text{host\_score}
\end{align}

Here, the final score combines the highest cosine similarity score from matched content chunks within each document, boosted BM25 scores, and a host-based weighting. Each document's chunked cosine score ensures finer-grained relevance, while BM25 and host boosts adjust for term frequency relevance and content source authority, respectively. This combination provides a more nuanced ranking by integrating multiple retrieval signals.

% \subsection{Retrieval Process}

% \begin{itemize}
%     \item \textbf{Initial Baseline (BM25 Only)}: We began with a BM25-only model as the base retrieval mechanism.
%     \item \textbf{Chunking + BM25}: To enhance accuracy, we incorporated chunking, allowing us to evaluate document segments separately and retrieve the most relevant content pieces.
%     \item \textbf{Chunking + BM25 + Host Boosting}: Finally, a host score was added to further differentiate the relevance based on document provenance, addressing cases where source reliability influences retrieval.
% \end{itemize}

\subsection{Parameter Selection}

The values for \texttt{bm25\_boost} and \texttt{host\_boost} were empirically determined by optimizing for the highest average similarity score between generated outputs and a predefined golden set. This iterative tuning process ensured that the scoring weights effectively balanced relevance and content source reliability. Table~\ref{tab:bm25_content_boost} and Table-~\ref{tab:host_boost} summarize how we select these parameters.

\subsection{Document Ranking}

After calculating scores for all documents, they were sorted in descending order, and the top 3 documents were returned as the final output, prioritizing the documents most likely to yield relevant information.

\section{Experiments and Evaluation}

\subsection{Datasets}
We evaluate our system using two carefully curated datasets: a golden dataset for measuring accuracy and performance, and a negative dataset for testing system robustness.

\subsubsection{Golden Dataset}
Our golden dataset comprises question-answer pairs from three primary documentation sources for our organization, with the following distribution:
\begin{itemize}
    \item creativecloud.adobe.com (Express Learn): 62 pairs
    \item www.adobe.com: 74 pairs
    \item helpx.adobe.com: 51 pairs
\end{itemize}

% \subsubsection{Golden Dataset}
% Our golden dataset comprises question-answer pairs from three primary documentation sources from the enterprise's product documentation, with the following distribution:
% \begin{itemize}
% \item Product Learning Portal: 62 pairs
% \item Main Corporate Website: 74 pairs
% \item Product Help Documentation: 51 pairs
% \end{itemize}

Each entry in the dataset contains:
\begin{itemize}
    \item A help query, initially LLM-generated and human-revised
    \item Relevant document URLs (separated by delimiters)
    \item Step-by-step answers, LLM-generated and human-revised
    \item Source documentation URL
\end{itemize}

\subsubsection{Negative Dataset}
To evaluate system robustness, we compiled a negative dataset consisting of:
\begin{itemize}
    \item 12 jailbreak attempts
    \item 6 NSFW queries
    \item 12 irrelevant queries
\end{itemize}
All queries were collected from internet sources and revised by human annotators, with an expected "content not found" response.

\subsection{Evaluation Setup}

Our evaluation framework uses the following common parameters across all experiments:
\begin{itemize}
    \item Generator Model: GPT-4o
    \item Retrieved URLs Limit: 3
\end{itemize}

\begin{figure}[!t]
    \centering
    \includegraphics[width=\columnwidth]{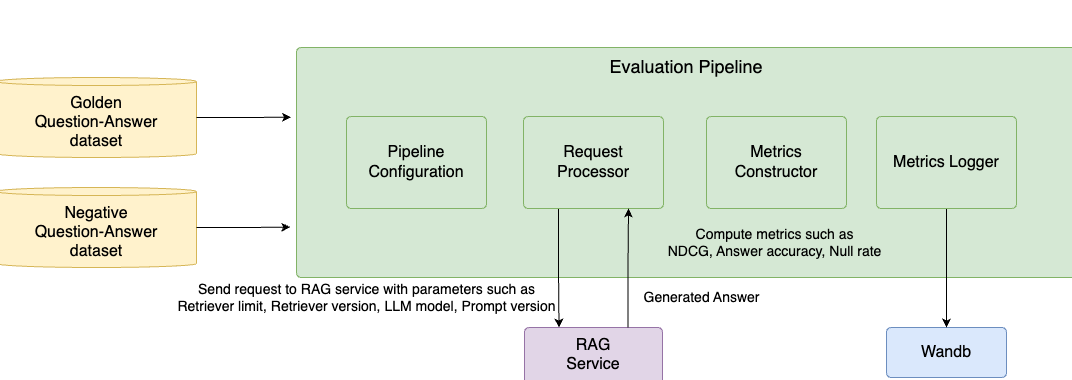}
    \caption{Evaluation Setup}
    \label{fig:eval_arch}
\end{figure}

\subsection{Results and Analysis}

\subsubsection{Effect of Chunk Size on Retrieval Performance}

To determine the optimal chunk size for document segmentation, we experimented with various chunk sizes and overlap values. Table~\ref{tab:chunk_size_comparison}
 summarizes the NDCG scores obtained for different configurations (see Appendix~\ref{app:metrics} for more details about the metric).

\begin{table}[H]
    \centering
    \resizebox{\columnwidth}{!}{  % Fit to column width
        \begin{tabular}{p{3.5cm}cccc}  % Reduced first column width
        \hline
        \textbf{Chunk size} & \textbf{Chunk Overlap} & \textbf{Context nDCG} \\
        \hline
        1000 & 100 & \textbf{0.828} \\
        2000 & 500 & 0.802 \\
        5000 & 1000 & 0.795 \\
        \hline
        \end{tabular}
    }
    \caption{Effect of Approximate Chunk Size on Retrieval Performance}
    \label{tab:chunk_size_comparison}
    \vspace{0.2cm}  % Add space after table
\end{table}

The results indicate that an approximate chunk size of 1000 characters with a split ensuring it ends on a sentence delimiter, with an overlap of 100 characters provides a favorable balance between retrieval granularity and computational efficiency.

\subsubsection{Boost Parameter Tuning}

We performed empirical tuning of the boost parameters to optimize the scoring algorithm for document retrieval.

\paragraph{BM25 Boost Tuning}

Using 60\% of the golden dataset as a validation set, we adjusted the BM25 content match boost. Table~\ref{tab:bm25_content_boost} shows the Top-3 NDCG scores for different boost values.

\begin{table}[H]
    \centering
    \resizebox{\columnwidth}{!}{  % Fit to column width
        \begin{tabular}{p{3.5cm}cccc}  % Reduced first column width
        \hline
        \textbf{bm25-boost} & \textbf{Context nDCG} \\
        \hline
        0.1  & 0.863 \\
        0.3  & \textbf{0.868} \\
        0.6  & 0.840 \\
        1  & 0.831 \\
        \hline
        \end{tabular}
    }
    \caption{Choosing optimal parameter for BM25}
    \label{tab:bm25_content_boost}
    \vspace{0.2cm}  % Add space after table
\end{table}

The optimal BM25 content boost was found to be 0.3, achieving the highest NDCG score.

\paragraph{Host Boost Tuning}

We also tuned the host boost parameter to enhance retrieval from authoritative sources. Table~\ref{tab:host_boost} presents the NDCG scores for varying host boost values.

\begin{table}[H]
    \centering
    \resizebox{\columnwidth}{!}{  % Fit to column width
        \begin{tabular}{p{3.5cm}cccc}  % Reduced first column width
        \hline
        \textbf{host-boost} & \textbf{Context nDCG} \\
        \hline
        0.1  & \textbf{0.853} \\
        0.3  & 0.842 \\
        0.6  & 0.830 \\
        1  & 0.781 \\
        \hline
        \end{tabular}
    }
    \caption{Choosing optimal parameter for BM25}
    \label{tab:host_boost}
    \vspace{0.2cm}  % Add space after table
\end{table}
A host boost value of 0.1 provided the best trade-off between relevance and authority.

\paragraph{Discussion of Boost Tuning Results}

The tuning of boost parameters significantly impacted retrieval performance. Higher boost values for BM25 content match improved NDCG up to a point, after which performance declined due to overemphasis on term frequency. Similarly, moderate host boosting enhanced the retrieval of authoritative documents without overshadowing relevance.

\subsubsection{Retrieval Strategy}

We evaluated five increasingly sophisticated retrieval strategies to assess their effectiveness in our domain-specific question-answering system. The strategies are:

\begin{itemize}

\item \textbf{Keyword Search (BM25)}: This baseline approach utilizes the BM25 algorithm for sparse keyword-based retrieval. It serves as a foundational benchmark for comparison.

\item \textbf{OpenAI text-embedding-3-large}: This strategy employs OpenAI's pre-trained text-embedding-3-large model for dense retrieval. It leverages semantic embeddings to capture the meaning of queries and documents without domain-specific fine-tuning.

\item \textbf{Fine-tuned Retriever }: As shown in Appendix~\ref{app:finetuning} this approach uses a dense retriever that has been fine-tuned on our domain-specific dataset, as discussed briefly in the Methods section. The fine-tuning enhances the model's ability to understand domain-specific terminology and nuances.

\item \textbf{Fine-tuned Retriever + Keyword Search}: This hybrid method combines our fine-tuned dense retriever with the BM25 keyword search. The integration aims to leverage both semantic understanding and exact keyword matching to improve retrieval performance.

\item \textbf{Fine-tuned Retriever + Keyword Search + Host Boost} Building upon the hybrid approach, this strategy incorporates URL host boosting. By assigning additional weight to documents from preferred URL hosts, we aim to enhance the relevance of the retrieved results in the enterprise context.

\end{itemize}

%% TODO:
%% We need the following structure
%% Only BM25, Retriever + BM25, Retriever + BM25 + host boost, then 
\begin{table}[H]
    \centering
    \resizebox{\columnwidth}{!}{  % Fit to column width
        \begin{tabular}{p{3.5cm}cccc}  % Reduced first column width
        \hline
        \textbf{Strategy} & \textbf{Context NDCG} \\
        \hline
        Keyword Search (BM25) & 0.640 \\
        text-embedding-3-large (openai)	& 0.760 \\
        Fine tuned retriever & 0.828 \\
        Fine tuned retriever + Keyword Search & 0.845 \\
        Fine tuned retriever + Keyword Search + Host boosting & \textbf{0.847} \\
        \hline
        \end{tabular}
    }
    \caption{Detailed Performance Comparison of Retrieval Strategies}
    \label{tab:retrieval-comparison}
    \vspace{0.2cm}  % Add space after table
\end{table}

\subsubsection{Guardrails around generated answer}
In order to protect any jailbreak attempts, we have implemented a guardrail mechanism on top of answer generation. Before returning the generated answer to the user, we compare the similarity of the generated answer with the system prompt, excluding the user query. If this similarity is very high, we do not return any content to the user. We have added jailbreak attempt queries in the evaluation dataset as well, which proves that we have made the system robust with this additional check.

\subsubsection{Generated Answer Analysis}

To validate our hypothesis that a better retriever leads to better final responses, we conducted an analysis comparing the generated answers from the LLM to the golden answers using an LLM as an evaluator.

\begin{table*}[h!]
    \centering
    \resizebox{\textwidth}{!}{  % Fit to column width
        \begin{tabular}{p{3.5cm}cccc}  % Reduced first column width
        \hline
        \textbf{Strategy} & \textbf{Answer Similarity mean} & \textbf{Answer Similarity Std} & \textbf{Groundedness mean} & \textbf{Groundedness Std}\\
        \hline
        Keyword Search (BM25) & 0.717 & 0.27 & 0.919 & 0.27\\
        text-embedding-3-large (openai)	& 0.748 & 0.23 & 0.979 & 0.13\\
        Fine tuned retriever & 0.755 & 0.22 & 0.974 & 0.15\\
        Fine tuned retriever + Keyword Search & 0.767 & 0.20 & 0.983 & 0.12\\
        Fine tuned retriever + Keyword Search + Host boosting & \textbf{0.780} & 0.19 & 0.983 & 0.12\\
        \hline
        \end{tabular}
    }
    \caption{Performance Comparison of Final answer with different retrieval strategies}
    \label{tab:generated-answer}
    \vspace{0.2cm}  % Add space after table
\end{table*}

\paragraph{Methodology}

Using GPT-4 as the evaluation model, we assessed the quality of the generated answers across different retrieval strategies. We focused on two key metrics:

Answer Similarity: Measures the degree of correspondence between the generated answer and the golden answer (see Appendix~\ref{app:metrics}).
Groundedness: Evaluates how well the generated answer is supported by the retrieved documents, ensuring factual correctness and context relevance.
For each query in the golden dataset, we generated answers using the LLM with contexts retrieved by each strategy. We then prompted GPT-4 to rate the similarity and groundedness of each generated answer compared to the golden answer on a scale from 0 to 1 (see Appendix~\ref{app:metrics}).

\paragraph{Results}

Table~\ref{tab:generated-answer}
presents the mean and std for answer similarity and groundedness across the different retrieval strategies.

The results indicate that strategies incorporating both the fine-tuned retriever and keyword search, especially with host boosting, achieved the highest scores in both answer similarity and groundedness.

\paragraph{Analysis}

The incremental improvements in both answer similarity and groundedness scores align with our hypothesis that enhanced retrieval strategies lead to better final responses. Specifically:

\begin{itemize}
    \item \textbf{Fine-tuned Retriever}: By capturing domain-specific semantics, it provides more relevant context to the LLM, improving answer quality.
    \item \textbf{Hybrid Approach}: Combining dense retrieval with keyword search leverages both semantic understanding and exact term matching, further enhancing the context.
    \item \textbf{Host Boosting}: Prioritizing authoritative sources increases the reliability of the information provided, which is reflected in higher groundedness scores.
\end{itemize}

\paragraph{Human-GPT Score Correlation Analysis}

To validate our LLM-based evaluation approach, we performed a focused correlation analysis using a subset of 14 queries that appeared in both golden dataset and a larger human evaluation dataset (containing approximately 200 queries). For these overlapping queries, we compared GPT-4's answer accuracy scores with binary human feedback (1 for correct, 0 for incorrect). While this represents a limited sample, it provides valuable initial insights into the alignment between automated and human evaluation methods. Analysis of these scores revealed:

\begin{itemize}
    \item Strong positive correlation between GPT-4 and human evaluations (Pearson correlation coefficient = $0.89$)
    \item GPT-4's scores showed good discrimination, with higher scores ($\mu = 0.82$) for queries rated correct by humans
    \item Only one case of significant disagreement occurred (for the Generative Fill query)
\end{itemize}

While this analysis is based on a small overlapping set, the strong correlation suggests promise in using LLM-based evaluation as a complement to human assessment. Future work should focus on expanding this comparison using a larger set of consistently evaluated queries to more comprehensively validate the automated evaluation approach.

These findings demonstrate that optimizing retrieval not only improves retrieval metrics but also has a significant positive impact on the quality of the final answers generated by the LLM.

\subsection{Discussion}
Our comprehensive evaluation of the hybrid search framework reveals several key findings and implications for domain-specific question answering systems:

\subsubsection{Performance Advantages of Hybrid Retrieval}

The experimental results demonstrate that our hybrid approach, combining fine-tuned dense retrieval with sparse search and host boosting, consistently outperforms single-method approaches across multiple metrics. This superior performance can be attributed to several factors:

\begin{itemize}
    \item Complementary Retrieval Signals: The integration of dense and sparse retrieval methods allows the system to leverage both semantic understanding and exact keyword matching. While the fine-tuned dense retriever captures domain-specific context and terminology, the BM25-based keyword search ensures critical term matches aren't overlooked.
    \item Host Boosting Effect: The addition of host-based boosting further refined results by prioritizing authoritative sources, leading to a modest but consistent improvement in both retrieval accuracy (NDCG: 0.84701) and answer quality metrics (Answer Similarity: 0.78, Groundedness: 0.983).
\end{itemize}

\subsubsection{Impact on Answer Generation}
The improvement in retrieval quality shows a direct correlation with enhanced answer generation:

\begin{itemize}
    \item \textbf{Answer Similarity}: The progression from basic keyword search (0.717) to the full hybrid approach (0.78) demonstrates that better context retrieval leads to more accurate answers.
    \item \textbf{Groundedness}: High groundedness scores across all methods indicate that the system maintains strong factual alignment, with the hybrid approach achieving the highest score (0.983).
\end{itemize}

\subsubsection{System Robustness}

Our evaluation framework's inclusion of negative test cases (jailbreak attempts, NSFW queries, and irrelevant queries) demonstrates the system's resilience to potential misuse. Detailed analysis of system performance against potentially problematic queries revealed:

\begin{table}[h]
\centering
\begin{tabular}{lll}
\hline
\textbf{Query Type} & \textbf{Performance} & \textbf{Details} \\
\hline
Jailbreak Attempts & 91.7\% & 11/12 null responses \\
NSFW Content & 100\% & 6/6 null responses \\
Irrelevant Queries & 100\% & 12/12 null responses \\
\hline
\end{tabular}
\caption{System Performance on Negative Queries}
\label{tab:negative_performance}
\end{table}

The system demonstrated particularly strong performance in handling NSFW content and irrelevant queries, with perfect accuracy in both categories. For jailbreak attempts, only one query out of twelve received any response, demonstrating robust resilience against prompt injection attempts. The implemented guardrail mechanism, which checks answer similarity against system prompts, provides an effective defense against jailbreak attempts while maintaining system usability.

\subsubsection{Practical Implications}

The findings have several important implications for enterprise deployments:

\begin{itemize}
    \item \textbf{Scalability}: The framework's architecture, built on Elasticsearch, provides a production-ready solution that can handle enterprise-scale deployments while maintaining performance.
    \item \textbf{Adaptability}: The tunable boost parameters (BM25\_boost and host\_boost) offer flexibility in adjusting the system for different domain-specific applications and content sources.
    \item \textbf{Cost-Effectiveness}: The chunking optimization (optimal size: 1000 tokens with 100 token overlap) provides an effective balance between retrieval accuracy and computational efficiency.
\end{itemize}

These findings provide a foundation for future research in domain-specific question answering systems while offering practical insights for enterprise implementations.

\subsection{Future Work}

% Addressing these limitations suggests several promising directions for future research:

\paragraph{Comprehensive Human Evaluation Framework}
Building upon our initial correlation analysis between LLM and human evaluations, future work should prioritize:
\begin{itemize}
    \item Development of a standardized evaluation protocol with consistent query sets for both human and automated evaluation, clear criteria for human evaluators and structured feedback collection mechanisms
    \item Large-scale human evaluation studies including systematic assessment of system performance across diverse query types, evaluation by both domain experts and end-users, collection of qualitative feedback on answer quality and usefulness and analysis of user satisfaction across different use cases
\end{itemize}

    \paragraph{Enhanced Context Integration}: Developing mechanisms to capture and utilize user context, including:
    \begin{itemize}
        \item Real-time user location tracking within the product
        \item Platform and device detection
        \item Context-aware response generation
    \end{itemize}

    \paragraph{Multilingual Extension}: Expanding the system to support multiple languages through:
    \begin{itemize}
        \item Multilingual dense retriever training
        \item Cross-lingual document retrieval
        \item Language-specific response generation
    \end{itemize}

    \paragraph{Multimodal Enhancement}: Incorporating visual content through:
    \begin{itemize}
        \item Multimodal embedding models
        \item Visual content understanding and extraction
        \item Integration of visual elements in response generation
        \item Screenshot and image-aware context processing
    \end{itemize}

These improvements would significantly enhance the system's utility and applicability across diverse enterprise environments.
\label{app:qualitative}

% \subsection{Representative Examples}
% Table~\ref{tab:qual_examples} shows representative examples of retrieval results from our model compared to the baseline.

% \begin{table}[h]
% \centering
% \caption{Qualitative comparison of retrieval results}
% \label{tab:qual_examples}
% \begin{tabular}{p{2.5cm}|p{5cm}|p{5cm}}
% \hline
% \textbf{Query} & \textbf{Our Model's Top Result} & \textbf{Baseline's Top Result} \\
% \hline
% [Example Query 1] & [Retrieved passage with explanation of why it's relevant] & [Baseline's retrieved passage with analysis] \\
% \hline
% [Example Query 2] & [Retrieved passage with explanation of why it's relevant] & [Baseline's retrieved passage with analysis] \\
% \hline
% \end{tabular}
% \end{table}

%%TODO:

%TODO: Franck and David to Add finetuning section

%TODO: Dewang to add qualitative examples and nDCG score description

% showng some qualitative examples would be good - sanat

\bibliography{aaai25}

\begin{thebibliography}{10}
\providecommand{\natexlab}[1]{#1}

\bibitem[{Arivazhagan et~al.(2023)Arivazhagan, Liu, Qi, Chen, Wang, and Huang}]{arivazhagan2023hybrid}
Arivazhagan, M.~G.; Liu, L.; Qi, P.; Chen, X.; Wang, W.~Y.; and Huang, Z. 2023.
\newblock Hybrid Hierarchical Retrieval for Open-Domain Question Answering.
\newblock In \emph{Findings of the Association for Computational Linguistics: ACL 2023}, 10680--10689.

\bibitem[{Eppalapally et~al.(2024)Eppalapally, Dangi, Bhat, Gupta, Zhang, Agarwal, Bagga, Yoon, Lipka, Rossi, and Dernoncourt}]{eppalapally-etal-2024-kapqa}
Eppalapally, S.; Dangi, D.; Bhat, C.; Gupta, A.; Zhang, R.; Agarwal, S.; Bagga, K.; Yoon, S.; Lipka, N.; Rossi, R.; and Dernoncourt, F. 2024.
\newblock {K}a{PQA}: Knowledge-Augmented Product Question-Answering.
\newblock In Yu, W.; Shi, W.; Yasunaga, M.; Jiang, M.; Zhu, C.; Hajishirzi, H.; Zettlemoyer, L.; and Zhang, Z., eds., \emph{Proceedings of the 3rd Workshop on Knowledge Augmented Methods for NLP}, 15--29. Bangkok, Thailand: Association for Computational Linguistics.

\bibitem[{Lazaridou, Cancedda, and Baroni(2022)}]{lazaridou2022internet}
Lazaridou, A.; Cancedda, N.; and Baroni, M. 2022.
\newblock Internet-Augmented Language Models through Few-Shot Prompting for Open-Domain Question Answering.
\newblock In \emph{Proceedings of the 60th Annual Meeting of the Association for Computational Linguistics (Volume 1: Long Papers)}, 437--456.

\bibitem[{Lewis et~al.(2020)Lewis, Perez, Piktus, Petroni, Karpukhin, Goyal, K{\"u}ttler, Lewis, Yih, Rockt{\"a}schel, Riedel, and Kiela}]{lewis2020retrieval}
Lewis, P.; Perez, E.; Piktus, A.; Petroni, F.; Karpukhin, V.; Goyal, N.; K{\"u}ttler, H.; Lewis, M.; Yih, W.-t.; Rockt{\"a}schel, T.; Riedel, S.; and Kiela, D. 2020.
\newblock Retrieval-Augmented Generation for Knowledge-Intensive NLP Tasks.
\newblock \emph{Advances in Neural Information Processing Systems}, 33: 9459--9474.

\bibitem[{Li et~al.(2021)Li, Ng, Xu, Zhu, Wang, and Xiang}]{li2021dual}
Li, A.~H.; Ng, P.; Xu, P.; Zhu, H.; Wang, Z.; and Xiang, B. 2021.
\newblock Dual Reader-Parser on Hybrid Textual and Tabular Evidence for Open Domain Question Answering.
\newblock arXiv:2108.02866.

\bibitem[{Oord, Li, and Vinyals(2018)}]{oord2018representation}
Oord, A. v.~d.; Li, Y.; and Vinyals, O. 2018.
\newblock Representation learning with contrastive predictive coding.
\newblock \emph{arXiv preprint arXiv:1807.03748}.

\bibitem[{Reimers(2019)}]{reimers2019sentence}
Reimers, N. 2019.
\newblock Sentence-BERT: Sentence Embeddings using Siamese BERT-Networks.
\newblock In \emph{Proceedings of the 2019 Conference on Empirical Methods in Natural Language Processing}. Association for Computational Linguistics.

\bibitem[{Sharma et~al.(2024)Sharma, Yoon, Dernoncourt, Sultania, Bagga, Zhang, Bui, and Kotte}]{sharma2024retrieval}
Sharma, S.; Yoon, D.~S.; Dernoncourt, F.; Sultania, D.; Bagga, K.; Zhang, M.; Bui, T.; and Kotte, V. 2024.
\newblock Retrieval Augmented Generation for Domain-specific Question Answering.
\newblock arXiv:2404.14760.

\bibitem[{Truera(2023)}]{trulens}
Truera. 2023.
\newblock Trulens.
\newblock \url{https://github.com/truera/trulens}.

\bibitem[{Zhu et~al.(2023)Zhu, Zhang, Zhai, and Liu}]{zhu2023hybrid}
Zhu, W.; Zhang, X.; Zhai, Q.; and Liu, C. 2023.
\newblock A Hybrid Text Generation-Based Query Expansion Method for Open-Domain Question Answering.
\newblock \emph{Future Internet}, 15(5): 180.

\end{thebibliography}

\appendix

\section{Model Fine-tuning Details}
\label{app:finetuning}

\subsection{Retriever Model Configuration}
We build a text encoder that computes a vector representation of a sentence. The model is trained to map user queries and corresponding documents in similar latent spaces so that it can be used as a semantic retriever~\cite{reimers2019sentence}.

% \subsection{Training Process}
% We use user behavior data to train the retriever. The behavior data consists of a ``user query," and a list of ``document" that were shown to the user by the in-house system. Among the list of ``document" we have a signal that a user clicked one of the provided documents. As each document consists of ``title" and ``body text," we can formulate the training loss as follows:
% $
% \mathcal{L}_\maqthrm{total}~:=~\mathcal{L}_\mathrm{1} (f_\theta(\text{q}),f_\theta(\text{title}))~+~\mathcal{L}_\mathrm{1} (f_\theta(\text{q}), f_\theta(\text{body text})) 
% $,
% where q is user query, $f_\theta()$ is sentence encoder parameterized by $\theta$. $\mathcal{L}_\mathrm{1}$, represents the InfoNCE loss function that operates between images and text~\cite{oord2018representation}.

\subsection{Training Process}
We use user behavior data to train the retriever. The behavior data consists of a "user query" and a list of "documents" that were shown to the user by the in-house system. Among the list of "documents," we have a signal that a user clicked on one of the provided documents. Since each document consists of a "title" and "body text," we formulate the training loss as follows:

\[
L_\text{total} := L_1 \big(f_\theta(\text{q}), f_\theta(\text{title})\big) + L_1 \big(f_\theta(\text{q}), f_\theta(\text{body text})\big),
\]

where \(q\) is the user query, \(f_\theta()\) is the sentence encoder parameterized by \(\theta\), and \(L_1\) represents the InfoNCE loss function that operates between images and text~\cite{oord2018representation}.

\section{Evaluation Metrics Details}

\label{app:metrics}

\subsection{Normalized Discounted Cumulative Gain (nDCG)}
nDCG is a standard metric in information retrieval that measures ranking quality. It is calculated as:
\begin{equation}
    \text{nDCG@k} = \frac{\text{DCG@k}}{\text{IDCG@k}}
\end{equation}
where DCG@k is defined as:
\begin{equation}
    \text{DCG@k} = \sum_{i=1}^k \frac{2^{rel_i} - 1}{\log_2(i + 1)}
\end{equation}
Here:
\begin{itemize}
    \item $rel_i$ is the relevance score of the result at position $i$
    \item IDCG@k is the DCG@k value for the ideal ranking
    \item The score ranges from 0 to 1, with 1 being perfect ranking
\end{itemize}

\subsection{Interpretation of nDCG Scores}
This metric is particularly suitable for our evaluation because:
\begin{itemize}
    \item It accounts for both relevance and position in the ranking
    \item It is normalized, allowing comparison across queries
\end{itemize}

\subsection{LLM-based Evaluation Metrics}
In addition to nDCG, we employ GPT as an automated judge to evaluate two key aspects of the generated responses: groundedness and accuracy.

\subsubsection{Groundedness Score}
We evaluate how well the generated responses are grounded in the retrieved context using the following prompt structure \cite{trulens}:

\begin{promptbox}
\textbf{System:} You are an impartial groundedness judge. You will be given a context and a response. Your task is to determine how grounded the response is in the given context. A response is considered grounded if it is supported by and does not contradict the given context.

Rate the groundedness on a scale from 0 to 1 (0.0, 0.1, 0.2, 0.3, 0.4, 0.5, 0.6, 0.7, 0.8, 0.9, 1.0), where 0 is completely ungrounded and 1 is perfectly grounded.

\textbf{Context:} \${context}

\textbf{Response:} \${response}

\textbf{Groundedness Score:}
\end{promptbox}

\subsubsection{Accuracy Score}
We evaluate the accuracy of generated responses by comparing them with ground truth answers using the following prompt \cite{trulens}:

\begin{promptbox}
\textbf{System:} You will be given one Model\_answer and a Groundtruth\_answer for a question about \${product}. Your task is to rate the similarity of the two answers on one metric.

\textbf{Evaluation Criteria:}\\
Similarity (0.0, 0.1, 0.2, 0.3, 0.4, 0.5, 0.6, 0.7, 0.8, 0.9, 1.0) - Similarity of overall text and similarity of each step or multiple steps also need to be considered.

\textbf{Question:} \${question}

\textbf{Groundtruth\_answer:} \${ground\_truth}

\textbf{Model\_answer:} \${model\_answer}

\textbf{Evaluation Steps:}
1. Read the question carefully
2. Do not modify the Groundtruth\_answer and Model\_answer
3. Read the Groundtruth\_answer and identify the steps and order
4. Read the Model\_answer and assess similarity to Groundtruth
5. Assign similarity score (0.0 to 1.0)
6. Reply with the Similarity score only
\end{promptbox}

Both metrics output scores in the range [0,1], where:
\begin{itemize}
    \item Groundedness Score measures how well the response is supported by the retrieved context
    \item Accuracy Score measures the semantic and structural similarity between the generated response and the ground truth
\end{itemize}

This combination of retrieval-based (nDCG) and generation-based (Groundedness, Accuracy) metrics provides a comprehensive evaluation framework for our system's performance.

\end{document}